\documentclass{article}
\usepackage{ijcai26}

\usepackage{times}
\usepackage{soul}
\usepackage{url}
\usepackage[hidelinks]{hyperref}
\usepackage[utf8]{inputenc}
\usepackage[small]{caption}
\usepackage{graphicx}
\usepackage{amsmath}
\usepackage{amsthm}
\usepackage{booktabs}
\usepackage{algorithm}
\usepackage{algorithmic}
\usepackage[switch]{lineno}

\usepackage{amssymb}
\usepackage[table]{xcolor}
\usepackage{multirow}
\usepackage{mathtools}
\usepackage{siunitx}

\title{SPOTR: Spatio-temporal Pooling One-Token Reconstruction for Universal Physiological Signal Self-supervised Learning}

\author{
Yiyu Gui$^{1, 2}$\and
Mingzhi Chen$^{1,2}$\and
Yuesheng Zhu$^{2}$\and
Guibo Luo$^{2*}$\And
Yuchao Yang$^{1*}$\\
\affiliations
$^1$ New Cornerstone Science Laboratory, Guangdong Provincial Key Laboratory of In-Memory Computing Chips, Shenzhen Graduate School, Peking University, Shenzhen, China\\
$^2$ Guangdong Provincial Key Laboratory of Ultra High Definition Immersive Media Technology, Shenzhen Graduate School, Peking University, Shenzhen, China\\
\emails
\{luogb, yuchaoyang\}@pku.edu.cn
}

\begin{document}

\maketitle

\begingroup
\renewcommand{\thefootnote}{\fnsymbol{footnote}}
\footnotetext[1]{Corresponding authors.}
\endgroup

\begin{abstract}
Physiological signals such as EEG, ECG, and PPG are widely used in clinical monitoring. Recent self-supervised learning (SSL) methods offer an attractive way to leverage unlabeled recordings, yet they still fall short in practice. In particular, current SSL methods struggle across heterogeneous datasets, often distorting clinically meaningful structures or learning shortcuts from temporal and cross-channel redundancy. Consequently, existing SSL methods often deliver limited performance under linear probing, a lightweight adaptation setting that better matches real-world medical scenarios. Moreover, most Transformer-based SSL models encode a flattened spatiotemporal token sequence, incurring high computation and memory cost, and are typically developed within a single modality. To address these limitations, we present \textbf{SPOTR} (\textbf{S}patio-temporal \textbf{P}ooling \textbf{O}ne-\textbf{T}oken \textbf{R}econstruction), a compress-reconstruct pretraining framework that introduces a single-token global bottleneck for physiological signals. SPOTR compresses each waveform into a single-token representation and reconstructs the signal conditioned only on this representation. Meanwhile, SPOTR introduces an efficient spatio-temporal compaction module to reduce computation and memory cost. Pretrained on \textbf{20} datasets spanning EEG, iEEG, ECG, and PPG, SPOTR consistently outperforms the strongest baseline under linear probing, improving average AUC by \textbf{18.49\%}, \textbf{21.71\%}, \textbf{17.86\%}, and \textbf{4.64\%}, respectively. Compared with a representative general-purpose time-series foundation model, SPOTR achieves around \textbf{78\%} lower latency and \textbf{52\%} lower peak GPU memory on average. The code can be found at \href{https://github.com/5GYYYYY/SPOTR}{https://github.com/5GYYYYY/SPOTR}.

\end{abstract}

\section{Introduction}

Physiological signals such as electroencephalography (EEG), electrocardiography (ECG), and photoplethysmography (PPG) are typically recorded as multi-channel, high-temporal-resolution waveforms, providing essential evidence for characterizing individual physiological states and health variations. Deep learning models have achieved remarkable progress in automated biosignal analysis \cite{medformer,medgnn} and have been widely applied to sleep analysis \cite{sleep1_attnsleep,sleep2_sleepfm,sleep3_xsleepfusion}, EEG-based seizure detection 
\cite{seizure2,seizure3,ssddb}, ECG-driven arrhythmia detection 
\cite{arrhythmia1,arrhythmia3}, and PPG-based hypertension detection \cite{hypertension1,hypertension2}. Despite their effectiveness, supervised methods often rely on large-scale datasets with high-quality annotations and expert review, which are costly and difficult to scale. To reduce reliance on expensive expert annotations, self-supervised learning (SSL) has made substantial advances leveraging unlabeled biosignals. Current SSL techniques for physiological signals can be broadly grouped into two categories: contrastive learning and masked reconstruction. Contrastive SSL (C-SSL) learns discriminative representations by bringing positives closer while separating negatives \cite{biot,papagei,pulseppg}. Masked reconstruction SSL (MR-SSL), in contrast, learns representations by predicting masked signal segments (or tokens) from unmasked ones \cite{labram,cbramod,heartlang,stmem}.

\begin{figure}[H]
\begin{center}
\centerline{\includegraphics[width=\columnwidth]{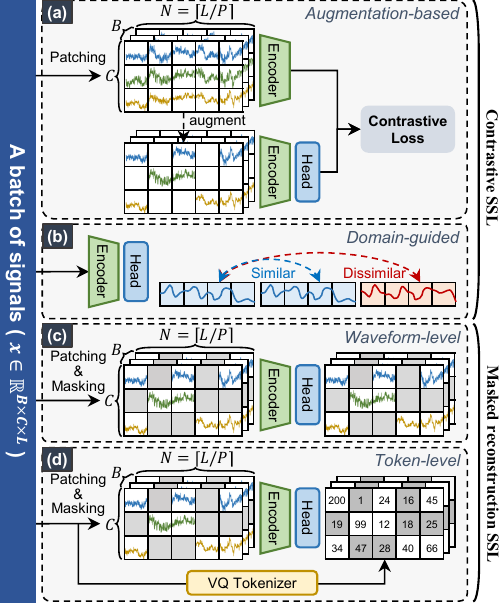}}
\caption{\textbf{Representative SSL paradigms for physiological signals.} (a) Augmentation-based contrastive learning (e.g., BIOT); (b) Domain-guided contrastive learning (e.g., PaPaGei, Pulse-PPG); (c) Waveform-level masked reconstruction (e.g., ST-MEM, CBraMod, CSBrain); (d) Token-level masked reconstruction (e.g., LaBraM, HeartLang).}

\label{fig1_baseline}
\end{center}
\end{figure}

\textbf{C-SSL: Performance Is Sensitive to View Design.}
Recent C-SSL methods mainly fall into two approaches: augmentation-based contrastive learning (e.g., BIOT \cite{biot}) and domain-guided contrastive learning (e.g., PaPaGei \cite{papagei} and Pulse-PPG \cite{pulseppg}). The former constructs multiple views via augmentations to encourage invariance in the learned representation, as shown in Fig. \ref{fig1_baseline}(a). The latter defines positive and negative pairs using task-specific or domain-specific rules and learns representations by pulling positives together while separating negatives, as shown in Fig. \ref{fig1_baseline}(b). While effective in specific settings, the learned invariances are highly dependent on pretraining design choices, and even slight changes in augmentation strength, view semantics, or pairing criteria can alter the supervision signal and induce unstable transfer behavior. In practice, this makes C-SSL harder to reliably generalize across heterogeneous datasets, where no single augmentation or pairing rule consistently preserves clinically relevant structures.

\textbf{MR-SSL: Masked Reconstruction Is Prone to Shortcut Learning.}
Current MR-SSL for biosignals includes reconstructing raw signals (e.g., ST-MEM \cite{stmem}, CBraMod \cite{cbramod}, and CSBrain \cite{csbrain}) and reconstructing discrete tokens (e.g., LaBraM \cite{labram} and HeartLang \cite{heartlang}). Their difference lies in whether the reconstruction target is continuous waveforms (Fig. \ref{fig1_baseline}(c)) or tokenized representations (Fig. \ref{fig1_baseline}(d)). However, under the mask-then-reconstruct setup, models can often exploit temporal continuity and cross-channel redundancy to recover missing content, creating shortcuts that bypass compact global representation learning. Consequently, the reconstruction objective can be weakly aligned with learning task-relevant, modality-stable structures, allowing pretraining loss to drop quickly while downstream performance improves only marginally.

\vspace{-0.5cm}

Taken together, these limitations leave key gaps in generalization for building foundation models. As a result, pretrained representations are often weak under \emph{linear probing}, a lightweight adaptation protocol that better matches real-world medical deployment where labeled data and compute are limited. In addition, existing architectures are frequently computationally heavy. As summarized in Table \ref{table1_baselines}, most Transformer-based SSL models process a flattened spatiotemporal sequence with length on the order of $C{\times}N$, where $C$ denotes the number of channels and $N$ the number of temporal tokens per channel, leading to high attention cost and memory usage. Finally, most previous studies remain modality-specific in practice, typically tailoring pretraining and evaluation to a single signal type rather than learning a generalizable representation across EEG, ECG, and PPG within one universal model.

\textbf{Our Approach.}
To address these gaps, we propose \textbf{SPOTR} (\textbf{S}patio-temporal \textbf{P}ooling \textbf{O}ne-\textbf{T}oken \textbf{R}econstruction), a universal SSL framework built around a \emph{compress–reconstruct} pretraining scheme with a \emph{single-token} global bottleneck. SPOTR compresses each waveform into a single-token representation and reconstructs the signal conditioned \emph{only} on this representation, suppressing shortcut learning and encouraging globally organized, generalizable features that benefit linear probing. To improve efficiency, SPOTR introduces an efficient spatiotemporal compaction module (ST Compactor) that reduces the token sequence length from $C \times N$ to $C{+}N$ before sequence modeling, substantially lowering computation while preserving spatiotemporal structure. Finally, by adopting a modality-agnostic objective and architecture, we enable unified pretraining across multiple physiological signal types, targeting a universal foundation model that can generalize across modalities and downstream datasets.

We summarize our main contributions as follows:

(1) We introduce SPOTR, a universal SSL framework that learns generalizable representations across physiological signal modalities via a compress–reconstruct pretraining scheme.

(2) We design an efficient spatiotemporal compaction architecture that shortens the encoder token sequence from $C\times N$ to $C{+}N$, improving training and inference efficiency without sacrificing representation quality.

(3) We demonstrate consistent gains across diverse downstream tasks and modalities, supporting SPOTR as a universal pretraining framework for physiological signals.

\section{Method}

\begin{figure*}[htb]
\begin{center}
\centerline{\includegraphics[width=\textwidth]{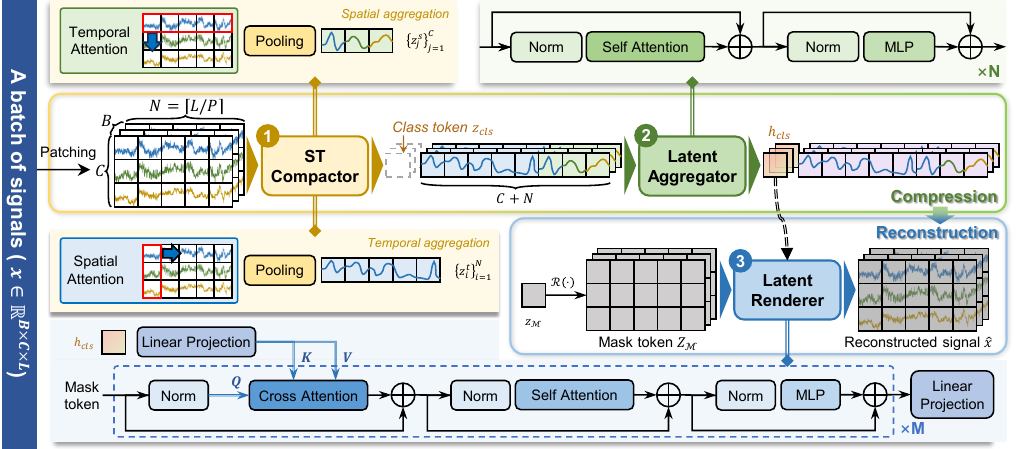}}
\caption{\textbf{Overview of SPOTR.} SPOTR performs compress--reconstruct pretraining with a single-token information bottleneck. The ST Compactor (1) compresses an input waveform into compact temporal tokens and spatial tokens. The Latent Aggregator (2) then fuses both streams into one global class token. For reconstruction, the Latent Renderer (3) starts from mask tokens and conditions the decoder on this single global token through cross-attention, forcing reconstruction to rely on global information rather than local visible context.}
\label{fig2_ourmodel}
\end{center}
\vspace{-0.5cm}
\end{figure*}

We present \textbf{SPOTR} (\textbf{S}patio-temporal \textbf{P}ooling \textbf{O}ne-\textbf{T}oken \textbf{R}econstruction), a compress-reconstruct pretraining framework that introduces a single-token global bottleneck for physiological signals. SPOTR comprises an \textbf{ST Compactor} $\mathcal{S}$ for spatiotemporal token compaction, a \textbf{Latent Aggregator} $\mathcal{A}$ for single-token global fusion, and a \textbf{Latent Renderer} $\mathcal{G}$ for bottleneck-conditioned signal reconstruction (Fig.~\ref{fig2_ourmodel}).

\subsection{SPOTR Training Objective}

Physiological signals often exhibit substantial redundancy across both time and channel dimensions. Under masked reconstruction, a decoder may still leverage visible local tokens or neighboring context to predict masked regions, so the reconstruction loss can be dominated by local correlations and common-mode structure—yielding shortcut learning that bypass global representation learning \cite{mae,clmae,diffmae}.

To address this issue and promote a more compact latent representation, SPOTR adopts a \textbf{compress–reconstruct pretraining scheme} with an explicit information bottleneck at both the encoder and decoder. The input is first compressed into two short sequences, then aggregated into a single class token as the global representation; reconstruction is conditioned only on this global token.

Let the physiological signal be $x \in \mathbb{R}^{B \times C \times L}$, where $B$ is the batch size, $C$ is the number of channels, and $L$ is the number of sample points. We denote the ST Compactor, Latent Aggregator, and Latent Renderer by $\mathcal{S}(\cdot)$, $\mathcal{A}(\cdot)$, and $\mathcal{G}(\cdot)$, respectively. With patch size $P$, we define the number of temporal tokens as $N=\left\lceil\frac{L}{P}\right\rceil$.

\textbf{(i) Spatiotemporal Compaction.}
The ST Compactor compresses the input into temporal tokens and spatial tokens, and uses learnable 2D positional encodings ${PE}_{2D}^\mathcal{A}$ to mark the relative position of each channel–time token:
\begin{equation}\label{eq1}
\begin{split}
Z=\mathcal{S}(x)+{PE}_{2D}^\mathcal{A}
&=[z_1^t,\ z_2^t,...,\ z_N^t;z_1^s,\ z_2^s,...,\ z_C^s] \\
&\in \mathbb{R}^{B \times (N+C) \times D_\mathcal{A}}
\end{split}
\end{equation}    
where $D_\mathcal{A}$ is the hidden dimension of the Latent Aggregator, $z_i^t$ is the $i$-th temporal token, and $z_j^s$ is the $j$-th spatial token.

\textbf{(ii) Single-token Global Aggregation.}
We add a learnable class token $z_{cls}$ and take the first output position as the global representation:
\begin{equation}\label{eq2}
h_{\mathrm{cls}} = A\big([z_{\mathrm{cls}}; Z]\big)[:,0:1] \in \mathbb{R}^{B \times 1 \times D_\mathcal{A}}
\end{equation}

Here $h_{cls}$ serves as the only information bottleneck in SPOTR, responsible for capturing global structure across both time and channels. With this design, the reconstruction is no longer conditioned on a large set of visible local tokens; instead, it relies solely on a fixed-size global representation, imposing a stronger constraint on representation compression and information organization.

\textbf{(iii) Mask-token Sequence and Conditional Rendering.} To reconstruct the signal, we build a mask-token sequence aligned with the target spatiotemporal grid. Let $z_\mathcal{M} \in \mathbb{R}^{1 \times 1 \times D_\mathcal{G}}$ be a learnable mask token, and let $\mathcal{R}(\cdot)$ be a repeat operator that expands a single token to $B\times C\times N$ positions and flattens them into a sequence:
\begin{equation}\label{eq3}
Z_\mathcal{M} = \mathcal{R}(z_\mathcal{M}; B, C, N) + \mathrm{PE}^{\mathcal{G}}_{2\mathrm{D}}
\in \mathbb{R}^{B \times (CN) \times D_\mathcal{G}}
\end{equation}
where $D_\mathcal{G}$ is the hidden dimension of the Latent Renderer and ${PE}_{2D}^\mathcal{G}$ is the decoder-side 2D positional encoding. The renderer conditions on $h_{cls}$ and outputs the reconstruction:
\begin{equation}\label{eq4}
\hat{x} = \mathcal{G}(Z_\mathcal{M}, h_{\mathrm{cls}}) \in \mathbb{R}^{B \times C \times L}
\end{equation}

\textbf{(iv) Reconstruction Loss.} Pretraining is optimized by minimizing the reconstruction error, and the objective can be formulated as the following empirical risk minimization problem:
\begin{equation}\label{eq5}
\mathcal{L}_{\mathrm{rec}} =
\mathbb{E}_{x \sim p_{\mathrm{data}}}\Big[\lVert \hat{x} - x \rVert_2^2\Big]
\end{equation}
where $p_{data}$ is the pretraining data distribution and $\hat{x}$ is the reconstructed signal produced end-to-end by the ST Compactor, Latent Aggregator, and Latent Renderer. By minimizing the reconstruction loss under a single-token information bottleneck, this objective drives the model to learn global representations that are both compact and reconstructive. Unlike contrastive learning, which relies on constructing positive and negative pairs to form training signals, the reconstruction objective uses the input itself as supervision, thereby avoiding the uncertainty introduced by negative-sample selection \cite{diffmae}.

\subsection{ST Compactor: Compressing Spatiotemporal Structure}

The ST Compactor $\mathcal{S}(\cdot)$ consists of a channel-independent convolutional patch embedding layer $\mathcal{C}$ followed by two orthogonal attention-based aggregation paths. The input signal $x \in \mathbb{R}^{B \times C \times L}$ is first segmented into temporal patches of length $P$ and projected into patch tokens:
\begin{equation}\label{eq6}
U = \mathcal{C}(x) \in \mathbb{R}^{B \times C \times N \times D_\mathcal{A}}
\end{equation}

Two attention operations are then applied: \paragraph{(i) Spatial Token Aggregation.} For each channel $j$, temporal self-attention is applied over its $N$ patches, followed by temporal pooling:
\begin{equation}\label{eq7}
z_j^s={pool}_t({MHA}_t(U_{:,j,:,:}))\in\mathbb{R}^{B\times D_\mathcal{A}},\ j=1,2,...,C
\end{equation}

\paragraph{(ii) Temporal Token Aggregation.} For each temporal token $i$, channel-wise self-attention aggregates information across channels, followed by spatial pooling:
\begin{equation}\label{eq8}
z_i^t={pool}_s({MHA}_s(U_{:,:,i,:}))\in\mathbb{R}^{B\times D_\mathcal{A}},\ i=1,2,...,N
\end{equation}

This design reduces the subsequent sequence length from $C \times N$ to $C+N$, preserving spatiotemporal structure while significantly lowering computation.

\subsection{Latent Aggregator: Aggregating into a Single Token}

The Latent Aggregator $\mathcal{A}(\cdot)$ is stacked Transformer encoder blocks that jointly encode $Z$ and aggregate it into the class token, yielding $h_{cls}$. By using the class token as the sole global representation throughout the network, $\mathcal{A}$ enforces a persistent single-token bottleneck, encouraging compact and transferable representations.

We adopt a pre-norm design with RMSNorm for training stability. The feed-forward network uses SwiGLU:
\begin{equation}\label{eq9}
{FFN}_{SwiGLU}(x)=(Swish(W_gx)\odot(W_1x))W_2
\end{equation}
where $\odot$ denotes the Hadamard product and  $W_g, W_1,W_2$ are learnable projection matrices. This design effectively enhances the model's ability to conduct feature selection and knowledge encoding under limited parameters.

\subsection{Latent Renderer: Rendering Signals from a Single Token Bottleneck}

The Latent Renderer $\mathcal{G}$ reconstructs the full physiological signal by conditioning on the highly compressed global latent $h_{cls}$ and progressively “rendering” the mask-token sequence $Z_\mathcal{M}$ into the signal domain. Concretely, $\mathcal{G}$ consists of an input projection $\mathcal{P}_{in}\in\mathbb{R}^{D_\mathcal{A}{\times D}_\mathcal{G}}$, an output projection $\mathcal{P}_{out}\in\mathbb{R}^{D_\mathcal{G}\times P}$, and a stack of Transformer decoder blocks. Here, $\mathcal{P}_{in}$ maps the conditioning vector to the decoder hidden space, while $\mathcal{P}_{out}$ regresses each token representation to a length-$P$ local segment, which is then reassembled to form the reconstructed signal.

Different from standard Transformer decoders that apply self-attention before cross-attention, we adopt a cross-attention-first update order, so that each position reads from the global condition $h_{cls}$ prior to any intra-sequence interaction. Specifically, let the input to the $k$-th decoder block be $Q^{(k-1)}\in\mathbb{R}^{B\times(CN)\times D_\mathcal{G}}$, $\widetilde{h}=\mathcal{P}_{in}(h_{cls})\in\mathbb{R}^{B\times1\times D_\mathcal{G}}$. We first inject the global condition via cross-attention:
\begin{equation}\label{eq10}
{\widetilde{Q}}^{(k)}=Q^{(k-1)}+MHA(Q^{(k-1)},\widetilde{h},\widetilde{h})
\end{equation}

Next, we perform self-attention over the mask-token sequence to enable information exchange across positions:
\begin{equation}\label{eq11}
{\bar{Q}}^{(k)}={\widetilde{Q}}^{(k)}+MHA({\widetilde{Q}}^{(k)},{\widetilde{Q}}^{(k)},{\widetilde{Q}}^{(k)})
\end{equation}

Finally, the resulting features are passed through a feed-forward network with a residual connection:
\begin{equation}\label{eq12}
Q^{(k)}={\bar{Q}}^{(k)}+{FFN}_{SwiGLU}({\bar{Q}}^{(k)})
\end{equation}

After stacking $K$ decoder blocks, we obtain $Q^{(K)}$. The final reconstruction is produced by projecting each position back to a length-$P$ segment and reshaping tokens into the original spatiotemporal layout:
\begin{equation}\label{eq13}
\hat{x}=Reshape(\mathcal{P}_{out}(Q^{(K)}))\in\mathbb{R}^{B\times C\times L}
\end{equation}

This design forces all local reconstructions to be mediated by the same global condition $h_{cls}$, thereby preventing the mask tokens from completing reconstruction solely via local correlations.

\section{Experimental Setup}

We pretrain and evaluate models under the SPOTR framework on multi-modal physiological signal datasets, and refer to the resulting family of models as \textbf{SPOTR}.

\begin{table*}[htb]
\caption{\textbf{Overview of baselines.} * \textit{NA indicates that the ``Number of tokens'' column is not applicable or non-Transformer architectures.}}
\label{table1_baselines}
\resizebox{\textwidth}{!}{%
\begin{tabular}{@{}c|cccc|cc|ccc@{}}
\toprule
                                & \multicolumn{4}{c|}{Signal Type} & \multicolumn{2}{c|}{}                                       &                                      &                                    &                                            \\
\multirow{-2}{*}{\textbf{Model}} & iEEG      & EEG       & ECG      & PPG      & \multicolumn{2}{c|}{\multirow{-2}{*}{Pretraining Paradigm}} & \multirow{-2}{*}{Model Architecture} & \multirow{-2}{*}{Number of tokens} & \multirow{-2}{*}{Adapt to any number of channels} \\ \midrule
\textbf{LaBraM}   (ICLR \cite{labram})              & $\checkmark$             & $\checkmark$         &          &          & Fig. \ref{fig1_baseline}(d)                        & MR-SSL                        & Transformer                          & C$\times$N                                &    $\checkmark$                                        \\
\textbf{ST-MEM}   (ICLR \cite{stmem})                       &              &           & $\checkmark$        &          & Fig. \ref{fig1_baseline}(c)                        & MR-SSL                        & Transformer                          & C$\times$N (C=12)                         &                                           \\
\textbf{MOMENT}   (ICML \cite{moment})                  & $\checkmark$          & $\checkmark$         & $\checkmark$        & $\checkmark$        & Fig. \ref{fig1_baseline}(c)                        & MR-SSL                        & Transformer                          & C$\times$N                                &    $\checkmark$                                        \\
\textbf{CBraMod}   (ICLR \cite{cbramod})            & $\checkmark$             & $\checkmark$         &          &          & Fig. \ref{fig1_baseline}(c)                        & MR-SSL                        & Transformer                          & C$\times$N                                &   $\checkmark$                                    \\
\textbf{HeartLang}   (ICLR \cite{heartlang})                 &                 &           & $\checkmark$        &          & Fig. \ref{fig1_baseline}(d)                        & MR-SSL                        & Transformer                          & C$\times$N (C=12)                         &                                           \\
\textbf{PaPaGei}   (ICLR \cite{papagei})                     &               &           &          & $\checkmark$        & Fig. \ref{fig1_baseline}(b)                        & C-SSL                         & ResNet                               & NA*                                &                                           \\
\textbf{Pulse-PPG}   (UBiComp \cite{pulseppg})                &               &           &          & $\checkmark$        & Fig. \ref{fig1_baseline}(b)                        & C-SSL                         & ResNet                               & NA*                                &                                           \\
\textbf{CSBrain}   (NeurIPS \cite{csbrain})              & $\checkmark$           & $\checkmark$         &          &          & Fig. \ref{fig1_baseline}(c)                        & MR-SSL                        & Transformer                          & C$\times$N                                &  $\checkmark$                                  \\
\rowcolor[HTML]{EFEFEF}
\textbf{SPOTR} (ours)                 & $\checkmark$               & $\checkmark$         & $\checkmark$        & $\checkmark$        & Fig. \ref{fig2_ourmodel}                          &      SPOTR                         & Transformer                          & C+N                                &     $\checkmark$                                  \\ \bottomrule
\end{tabular}%
}
\end{table*}

\begin{table*}[!ht]
\caption{\textbf{Overview of downstream datasets.} ``\# Channels'' and ``\# Subjects'' denote the number of channels and subjects in the dataset.}
\label{table2_downstream_datasets}
\resizebox{\textwidth}{!}{%
\begin{tabular}{@{}c|c|c|cccc@{}}
\toprule
\textbf{Signal modality}  & \textbf{Dataset} & \textbf{Task}                            & \textbf{Label} & \textbf{\# Channels} & \textbf{Duration} & \textbf{\# Subjects} \\ \midrule
                                      & Mayo \cite{mayo}            & Intermittent epileptiform discharges (IEDs) detection      & 2-class        & 1                  & 3s                & 15                 \\
\multirow{-2}{*}{iEEG}                & FNUSA \cite{mayo}           & Intermittent epileptiform discharges (IEDs) detection      & 2-class        & 1                  & 3s                & 12                 \\ \midrule
                                      & ISRUC \cite{isruc}           & Sleep Stage Classification               & 5-class        & 2                  & 30s               & 100                \\
                                      & Siena \cite{siena}           & Seizure detection                        & 2-class        & 29                 & 4s                & 13                 \\
                                      & NTUH-BIS \cite{ntuh-bis}        & Anesthesia status detection              & 4-class        & 1                  & 30s               & 23                 \\
                                      & MDD \cite{mdd}             & Major depressive disorder detection      & 2-class        & 19                 & 5s                & 64                 \\
\multirow{-5}{*}{EEG}                 & Schizo-28 \cite{schizo}        & Schizophrenia detection                  & 2-class        & 19                 & 10s               & 28                 \\ \midrule
                                      & CPSC2018 \cite{cpsc2018}        & Heart disease detection                  & 9-class        & 12                 & 10s               & 6877               \\
                                      & PTB-sub \cite{ptbxl}         & Heart disease detection (Fine-grained)   & 23-class       & 12                 & 10s               & 18885              \\
                                      & PTB-super \cite{ptbxl}       & Heart disease detection (Coarse-grained) & 5-class        & 12                 & 10s               & 18885              \\
                                      & PTB-rhythm \cite{ptbxl}      & Heart disease detection (Rhythm)         & 12-class       & 12                 & 10s               & 18885              \\
\multirow{-5}{*}{ECG}                 & Europe ST-T \cite{europe_st-t}     & Heartbeat Classification                 & 5-class        & 2                  & 1s                & 90                 \\ \midrule
                                      & PPG-BP \cite{ppg-bp}          & Hypertension detection                   & 2-class        & 3                  & 2.1s              & 219                \\
                                      & PPG-Dalia \cite{ppgdalia}       & Human activity recognition               & 9-class        & 1                  & 8s                & 15                 \\
                                      & WESAD$_v$ \cite{wesad}         & Emotion classification                   & 2-class        & 1                  & 10s               & 15                 \\
                                      & WESAD$_a$ \cite{wesad}         & Emotion classification                   & 2-class        & 1                  & 10s               & 15                 \\
\multirow{-5}{*}{PPG}                 & CLBP \cite{clbp}            & Hypertension detection                   & 2-class        & 1                  & 10s               & 12                 \\ \bottomrule
\end{tabular}%
}
\vspace{-0.2cm}
\end{table*}

\subsection{Baselines}

As summarized in Table \ref{table1_baselines}, we compare \textbf{SPOTR} with representative SSL baselines. For each modality, we include (i) a general-purpose time-series SSL foundation model, \textbf{MOMENT}, and (ii) domain-specific SSL models. Specifically, for EEG/iEEG we use \textbf{LaBraM}, \textbf{CBraMod}, and \textbf{CSBrain}; for ECG we use \textbf{ST-MEM} and \textbf{HeartLang}; and for PPG we use \textbf{PaPaGei} and \textbf{Pulse-PPG}. We do not include \textbf{BIOT} \cite{biot} in our comparison because, although it claims applicability to multiple physiological modalities, only pretrained EEG weights are publicly available. More details are provided in the Appendix.

\subsection{Datasets}

\subsubsection{Pretraining Datasets}

We curate a large-scale pretraining dataset covering 4 physiological modalities, including scalp EEG, intracranial EEG (iEEG), ECG, and PPG. Overall, the pretraining data collection contains \textbf{20 datasets}, including \textbf{6} EEG datasets \cite{alzheimer,sleepedf,shhs,chbmit,tdbrain,nmt}, \textbf{2} iEEG datasets \cite{multicenterieeg,swecethz}, \textbf{8} ECG datasets \cite{mimic,shaoxing,shandong,ludb,ptbdiagnostic,stpeter,geogia,code15}, \textbf{3} PPG datasets \cite{pulsetransitppg,butppg,sdb}, and \textbf{one} multimodal dataset \cite{vitaldb}. In total, it provides \textbf{over 17 million signal samples} from \textbf{more than 450,000 subjects}, spanning diverse acquisition settings. More details are provided in the Appendix.

\subsubsection{Downstream Datasets}

As shown in Table \ref{table2_downstream_datasets}, we conduct a comprehensive evaluation of SPOTR on \textbf{17} downstream datasets spanning multiple physiological signal modalities, including iEEG, EEG, ECG, and PPG. For all the downstream datasets, we resampled the signals to 200 Hz and applied a 50/60 Hz notch filter (depending on the acquisition region) to suppress power-line interference. More details are provided in the Appendix.

\subsection{Implementation Details}

All experiments were conducted on a Linux server with an NVIDIA A100 80GB GPU, using PyTorch 2.0.1 and CUDA 12.2. For pretraining, we used AdamW with $\beta=(0.9,0.999)$, weight decay 0.1, and a peak learning rate of $2\times{10}^{-4}$. Each dataset was trained with a per-step batch size of 256 and gradient accumulation of 20, resulting in an effective batch size of 5120. We pretrained for 3 epochs with a learning-rate schedule including a 10\% warm-up; bfloat16 precision was used for computational efficiency.

For downstream evaluation, we used subject-independent train/validation/test splits for all tasks and report the mean and standard deviation of test-set performance over 5 independent runs with different random seeds. We kept the optimizer as AdamW with the same $\beta$ and weight decay, but used a constant learning rate of $2\times{10}^{-4}$ and a batch size of 128. All downstream runs were trained for up to 200 epochs with early stopping based on validation AUC (patience = 20 epochs). Unless otherwise specified, downstream training was performed in float32 for stable optimization.

\begin{table*}[ht]
\caption{\textbf{Linear-probing performance comparison.} \textbf{Bold} denotes the best result and \underline{underline} denotes the second best. For SPOTR, subscripts report the absolute AUC gap relative to the strongest baseline: $\uparrow$ denotes the improvement over the second-best model when SPOTR attains the top performance, whereas $\downarrow$ denotes the gap to the best-performing model otherwise.}
\vspace{-0.1cm}
\label{table3_result_linear}
\resizebox{\textwidth}{!}{%
\begin{tabular}{@{}ccccccccccccc@{}}
\toprule
\rowcolor{yellow!15}
\multicolumn{13}{c}{iEEG} \\
\multicolumn{2}{c|}{\textbf{Dataset}} & \multicolumn{5}{c}{\textbf{Mayo}} & \multicolumn{5}{c|}{\textbf{FNUSA}} & \\
Model & \multicolumn{1}{l|}{Applicable scope} & \multicolumn{3}{c}{mean} & \multicolumn{2}{c}{std.} & \multicolumn{2}{c}{mean} & \multicolumn{3}{c|}{std.} & \multirow{-2}{*}{Ave} \\ \midrule
MOMENT  & \multicolumn{1}{c|}{\textit{Time-series}}           & \multicolumn{3}{c}{50.46} & \multicolumn{2}{c}{3.29}  & \multicolumn{2}{c}{50.02} & \multicolumn{3}{c|}{2.83} & 50.24 \\
LaBraM & \multicolumn{1}{c|}{\textit{EEG/iEEG-only}}         & \multicolumn{3}{c}{{\ul{63.69}}} & \multicolumn{2}{c}{4.60} & \multicolumn{2}{c}{{\ul{69.32}}} & \multicolumn{3}{c|}{5.78} & {\ul{66.51}} \\
CBraMod & \multicolumn{1}{c|}{\textit{EEG/iEEG-only}}         & \multicolumn{3}{c}{46.65} & \multicolumn{2}{c}{17.67} & \multicolumn{2}{c}{62.83} & \multicolumn{3}{c|}{4.16} & 54.74 \\
CSBrain & \multicolumn{1}{c|}{\textit{EEG/iEEG-only}}         & \multicolumn{3}{c}{59.64} & \multicolumn{2}{c}{2.67}  & \multicolumn{2}{c}{55.91} & \multicolumn{3}{c|}{0.64} & 57.78 \\
\rowcolor[HTML]{EFEFEF}
SPOTR   & \multicolumn{1}{c|}{\textit{Physiological signals}} &
\multicolumn{3}{c}{$\mathbf{88.92}_{\textcolor{red}{\uparrow\,25.23}}$} &
\multicolumn{2}{c}{0.06} &
\multicolumn{2}{c}{$\mathbf{87.52}_{\textcolor{red}{\uparrow\,18.20}}$} &
\multicolumn{3}{c|}{0.16} &
$\mathbf{88.22}_{\textcolor{red}{\uparrow\,21.71}}$ \\ \midrule

\rowcolor{green!15}
\multicolumn{13}{c}{EEG} \\
\multicolumn{2}{c|}{\textbf{Dataset}} & \multicolumn{2}{c}{\textbf{ISRUC}} & \multicolumn{2}{c}{\textbf{Siena}} & \multicolumn{2}{c}{\textbf{NTUH-BIS}} & \multicolumn{2}{c}{\textbf{MDD}} & \multicolumn{2}{c|}{\textbf{Schizo-28}} & \\
Model & \multicolumn{1}{c|}{Applicable scope} & mean & std. & mean & std. & mean & std. & mean & std. & mean & \multicolumn{1}{c|}{std.} & \multirow{-2}{*}{Ave} \\ \midrule
MOMENT  & \multicolumn{1}{c|}{\textit{Time-series}}   & 71.16 & 0.89 & 62.45 & 5.17 & 55.09 & 2.17 & 79.13 & 4.25 & {\ul{68.73}} & \multicolumn{1}{c|}{8.85} & 67.31 \\
LaBraM & \multicolumn{1}{c|}{\textit{EEG/iEEG-only}} & {\ul{81.63}} & 2.43 & {\ul{66.87}} & 9.39 & {\ul{63.67}} & 4.87 & 81.23 & 4.15 & 53.58 & \multicolumn{1}{c|}{5.56} & 69.40 \\
CBraMod & \multicolumn{1}{c|}{\textit{EEG/iEEG-only}} & 77.20 & 1.22 & 58.25 & 2.86 & 54.53 & 1.59 & {\ul{90.03}} & 1.37 & 68.04 & \multicolumn{1}{c|}{6.16} & {\ul{69.61}} \\
CSBrain & \multicolumn{1}{c|}{\textit{EEG/iEEG-only}} & 60.57 & 2.27 & 55.68 & 3.38 & 54.75 & 1.26 & 78.37 & 3.26 & 59.47 & \multicolumn{1}{c|}{4.61} & 61.77 \\
\rowcolor[HTML]{EFEFEF}
SPOTR   & \multicolumn{1}{c|}{\textit{Physiological signals}} &
$\mathbf{92.83}_{\textcolor{red}{\uparrow\,11.21}}$ & 0.03 &
$\mathbf{76.87}_{\textcolor{red}{\uparrow\,10.00}}$ & 0.90 &
$\mathbf{87.69}_{\textcolor{red}{\uparrow\,24.02}}$ & 0.13 &
$\mathbf{95.31}_{\textcolor{red}{\uparrow\,5.28}}$  & 0.12 &
$\mathbf{87.79}_{\textcolor{red}{\uparrow\,19.06}}$ & \multicolumn{1}{c|}{8.87} &
$\mathbf{88.10}_{\textcolor{red}{\uparrow\,18.49}}$ \\ \midrule

\rowcolor{blue!15}
\multicolumn{13}{c}{ECG} \\
\multicolumn{2}{c|}{\textbf{Dataset}} & \multicolumn{2}{c}{\textbf{CPSC2018}} & \multicolumn{2}{c}{\textbf{PTB-sub}} & \multicolumn{2}{c}{\textbf{PTB-super}} & \multicolumn{2}{c}{\textbf{PTB-rhythm}} & \multicolumn{2}{c|}{\textbf{Europe ST-T}} & \\
Model & \multicolumn{1}{c|}{Applicable scope} & mean & std. & mean & std. & mean & std. & mean & std. & mean & \multicolumn{1}{c|}{std.} & \multirow{-2}{*}{Ave} \\ \midrule
MOMENT    & \multicolumn{1}{c|}{\textit{Time-series}} & 58.90 & 0.92 & 62.44 & 1.07 & 61.99 & 1.76 & 63.07 & 2.94 & {\ul{68.87}} & \multicolumn{1}{c|}{1.40} & 63.05 \\
ST-MEM    & \multicolumn{1}{c|}{\textit{ECG-only}}    & 53.56 & 0.38 & 56.44 & 1.25 & 56.98 & 1.40 & 55.89 & 0.96 & 51.02 & \multicolumn{1}{c|}{2.25} & 54.78 \\
HeartLang & \multicolumn{1}{c|}{\textit{ECG-only}}    & {\ul{69.87}} & 0.39 & {\ul{77.31}} & 0.28 & {\ul{77.44}} & 0.33 & {\ul{66.65}} & 0.40 & 56.38 & \multicolumn{1}{c|}{1.21} & {\ul{69.53}} \\
\rowcolor[HTML]{EFEFEF}
SPOTR     & \multicolumn{1}{c|}{\textit{Physiological signals}} &
$\mathbf{92.33}_{\textcolor{red}{\uparrow\,22.46}}$ & 0.03 &
$\mathbf{90.21}_{\textcolor{red}{\uparrow\,12.90}}$ & 0.02 &
$\mathbf{89.10}_{\textcolor{red}{\uparrow\,11.66}}$ & 0.01 &
$\mathbf{82.84}_{\textcolor{red}{\uparrow\,16.19}}$ & 0.04 &
$\mathbf{82.47}_{\textcolor{red}{\uparrow\,13.60}}$ & \multicolumn{1}{c|}{0.19} &
$\mathbf{87.39}_{\textcolor{red}{\uparrow\,17.86}}$ \\ \midrule

\rowcolor{purple!15}
\multicolumn{13}{c}{PPG} \\
\multicolumn{2}{c|}{\textbf{Dataset}} & \multicolumn{2}{c}{\textbf{PPG-BP}} & \multicolumn{2}{c}{\textbf{PPG-Dalia}} & \multicolumn{2}{c}{\textbf{WESAD$_{\text{v}}$}} & \multicolumn{2}{c}{\textbf{WESAD$_{\text{a}}$}} & \multicolumn{2}{c|}{\textbf{CLBP}} & \\
Model & \multicolumn{1}{c|}{Applicable scope} & mean & std. & mean & std. & mean & std. & mean & std. & mean & \multicolumn{1}{c|}{std.} & \multirow{-2}{*}{Ave} \\ \midrule
MOMENT    & \multicolumn{1}{c|}{\textit{Time-series}} & 47.14 & 14.74 & {\ul{68.07}} & 2.56 & {\ul{67.63}} & 2.47 & {\ul{61.68}} & 3.44 & {\ul{52.98}} & \multicolumn{1}{c|}{0.40} & {\ul{59.50}} \\
PaPaGei   & \multicolumn{1}{c|}{\textit{PPG-only}}    & \textbf{55.97} & 14.29 & 51.47 & 0.33 & 53.85 & 2.99 & 47.53 & 3.67 & 49.81 & \multicolumn{1}{c|}{0.22} & 51.73 \\
Pulse-PPG & \multicolumn{1}{c|}{\textit{PPG-only}}    & 46.58 & 13.24 & 57.28 & 1.37 & 49.31 & 8.17 & 45.88 & 3.96 & 51.85 & \multicolumn{1}{c|}{0.74} & 50.18 \\
\rowcolor[HTML]{EFEFEF}
SPOTR     & \multicolumn{1}{c|}{\textit{Physiological signals}} &
$\underline{51.99}_{\textcolor{green}{\downarrow\,3.98}}$ & 8.26 &
$\mathbf{82.12}_{\textcolor{red}{\uparrow\,14.05}}$ & 0.05 &
$\mathbf{69.03}_{\textcolor{red}{\uparrow\,1.40}}$ & 0.49 &
$\mathbf{62.06}_{\textcolor{red}{\uparrow\,0.38}}$ & 0.19 &
$\mathbf{55.52}_{\textcolor{red}{\uparrow\,2.54}}$ & \multicolumn{1}{c|}{0.25} &
$\mathbf{64.14}_{\textcolor{red}{\uparrow\,4.64}}$ \\
\bottomrule
\end{tabular}%
}
\end{table*}

\section{Results}

\begin{figure}[htb]
\vspace{-0.2cm}
\begin{center}
\centerline{\includegraphics[width=\columnwidth]{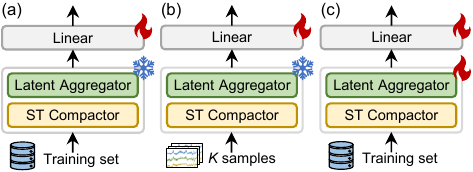}}
\caption{\textbf{Experimental setup.}}
\label{experiment_setup}
\end{center}
\vspace{-0.7cm}
\end{figure}

We evaluate three adaptation protocols: \textbf{(i) Linear probing}, where the pretrained backbone is frozen and only a single linear classification head is trained (Fig. \ref{experiment_setup}(a)); \textbf{(ii) Few-shot classification}, where we construct a support set by randomly sampling $K\in{1,2,4,8}$ labeled examples per class from the training set, freeze the backbone, and train only the same linear head (Fig. \ref{experiment_setup}(b)); and \textbf{(iii) Full fine-tuning}, where we train the backbone and the classification head under the same downstream configuration (Fig. \ref{experiment_setup}(c)). In all the experiments, we use AUC as the evaluation metric.

\vspace{-0.2cm}
\subsection{Linear Probing}

Table \ref{table3_result_linear} summarizes linear-probing performance, which directly reflects the quality of pretrained representations under lightweight adaptation.

On iEEG, SPOTR achieves the best performance with an average AUC of 88.22\%, surpassing the strongest iEEG/EEG SSL baseline by over 20\%. On EEG, SPOTR is consistently superior across the five datasets, reaching an average AUC of 88.10\% and exceeding EEG-specific SSL baselines. On ECG, SPOTR again delivers the strongest results, achieving an average AUC of 87.39\% and improving over the best ECG-specific SSL baseline by 17.86\%. On PPG, SPOTR attains the highest average AUC of 64.14\%, outperforming PPG-specific baselines (PaPaGei and Pulse-PPG) on most datasets. Since MOMENT is evaluated on all datasets, we further summarize results over all 17 datasets in Table \ref{table3_result_linear}. SPOTR achieves an overall mean AUC of \textbf{80.86\%}, exceeding MOMENT by \textbf{19.11\%}.

Across four physiological modalities, SPOTR consistently surpasses both the general time-series baseline and modality-specific SSL models in the linear-probing setting, supporting its practicality as a universal and lightweight-transfer pretraining framework for physiological signals.

\subsection{Few-shot Classification}

\begin{figure}[htb]
\begin{center}
\centerline{\includegraphics[width=0.95\columnwidth]{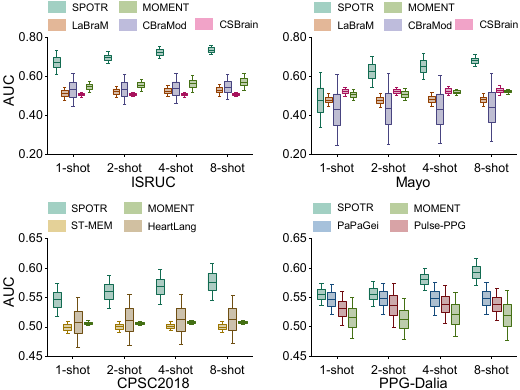}}
\vspace{-0.3cm}
\caption{\textbf{Few-shot classification results.} Boxes show the 25–75th percentiles with the median line; whiskers indicate the spread across repeated runs under each 1/2/4/8-shot setting.}
\label{fig3_few_shot}
\end{center}
\vspace{-0.4cm}
\end{figure}

As shown in Fig. \ref{fig3_few_shot}, under 1/2/4/8-shot settings, SPOTR achieves the best or near-best AUC on four representative datasets, each corresponding to a different modality: ISRUC (EEG), Mayo (iEEG), CPSC2018 (ECG), and PPG-Dalia (PPG). Across all modalities, SPOTR exhibits a stable and consistent improvement as more labeled examples are provided, indicating that it can make good use of very limited labeled data. Moreover, the relatively tight performance dispersion across runs suggests robust adaptation under limited supervision. Overall, these results support SPOTR for practical few-shot scenarios, where annotation is scarce but reliable adaptation is still needed.

\vspace{-0.25cm}

\subsection{Full Fine-tuning}

As shown in Table \ref{table4_full_fine-tune}, under the full-parameter fine-tuning setting, SPOTR remains competitive across all four modalities. It achieves the best average performance on iEEG (90.25\%) and EEG (89.61\%), and shows the clearest advantage on ECG with the highest mean score (92.70\%). On PPG, SPOTR is overall comparable to the strongest PPG-specific baseline, with only a marginal difference in average performance. Overall, even when end-to-end fine-tuning reduces the gaps among pretraining paradigms, SPOTR still provides robust transfer across modalities and datasets.

\begin{table}[htb]
\caption{\textbf{Full fine-tuning performance comparison.}}
\label{table4_full_fine-tune}
\centering
\resizebox{0.95\columnwidth}{!}{%
\begin{tabular}{@{}cccccc@{}}
\toprule
\rowcolor{yellow!15}
\multicolumn{6}{c}{iEEG} \\
\multicolumn{1}{c|}{\textbf{Dataset}} & \multicolumn{2}{c}{\textbf{Mayo}} & \multicolumn{2}{c|}{\textbf{FNUSA}} & \\
\multicolumn{1}{c|}{Model} & mean & std. & mean & \multicolumn{1}{c|}{std.} & \multirow{-2}{*}{Ave} \\ \midrule
\multicolumn{1}{c|}{MOMENT}  & \textbf{89.69} & 0.03 & {\ul{89.11}} & \multicolumn{1}{c|}{0.01} & {\ul{89.40}} \\
\multicolumn{1}{c|}{LaBraM} & 86.20 & 1.23 & 87.71 & \multicolumn{1}{c|}{1.58} & 86.96 \\
\multicolumn{1}{c|}{CBraMod} & 88.17 & 1.79 & 87.08 & \multicolumn{1}{c|}{2.23} & 87.63 \\
\multicolumn{1}{c|}{CSBrain} & 87.63 & 1.51 & 87.22 & \multicolumn{1}{c|}{9.56} & 87.43 \\
\rowcolor[HTML]{EFEFEF}
\multicolumn{1}{c|}{SPOTR} &
$\underline{89.28}_{\textcolor{green}{\downarrow\,0.41}}$ & 1.32 &
$\mathbf{91.21}_{\textcolor{red}{\uparrow\,2.10}}$ & \multicolumn{1}{c|}{0.81} &
$\mathbf{90.25}_{\textcolor{red}{\uparrow\,0.85}}$ \\ \midrule

\rowcolor{green!15}
\multicolumn{6}{c}{EEG} \\
\multicolumn{1}{c|}{\textbf{Dataset}} & \multicolumn{2}{c}{\textbf{ISRUC}} & \multicolumn{2}{c|}{\textbf{Siena}} & \\
\multicolumn{1}{c|}{Model} & mean & std. & mean & \multicolumn{1}{c|}{std.} & \multirow{-2}{*}{Ave} \\ \midrule
\multicolumn{1}{c|}{MOMENT}  & 87.91 & 0.01 & {\ul{83.29}} & \multicolumn{1}{c|}{1.13} & 85.60 \\
\multicolumn{1}{c|}{LaBraM} & 93.52 & 0.27 & 77.95 & \multicolumn{1}{c|}{1.20} & {\ul{85.74}} \\
\multicolumn{1}{c|}{CBraMod} & \textbf{96.98} & 0.23 & 74.04 & \multicolumn{1}{c|}{3.83} & 85.51 \\
\multicolumn{1}{c|}{CSBrain} & 93.46 & 0.20 & 57.82 & \multicolumn{1}{c|}{4.86} & 75.64 \\
\rowcolor[HTML]{EFEFEF}
\multicolumn{1}{c|}{SPOTR} &
$\underline{95.46}_{\textcolor{green}{\downarrow\,1.54}}$ & 0.16 &
$\mathbf{83.76}_{\textcolor{red}{\uparrow\,0.47}}$ & \multicolumn{1}{c|}{3.01} &
$\mathbf{89.61}_{\textcolor{red}{\uparrow\,3.87}}$ \\ \midrule

\rowcolor{blue!15}
\multicolumn{6}{c}{ECG} \\
\multicolumn{1}{c|}{\textbf{Dataset}} & \multicolumn{2}{c}{\textbf{CPSC2018}} & \multicolumn{2}{c|}{\textbf{Europe ST-T}} & \\
\multicolumn{1}{c|}{Model} & mean & std. & mean & \multicolumn{1}{c|}{std.} & \multirow{-2}{*}{Ave} \\ \midrule
\multicolumn{1}{c|}{MOMENT}    & 79.48 & 1.47 & {\ul{85.79}} & \multicolumn{1}{c|}{0.63} & {\ul{82.64}} \\
\multicolumn{1}{c|}{ST-MEM}    & 74.64 & 1.08 & 82.34 & \multicolumn{1}{c|}{2.18} & 78.49 \\
\multicolumn{1}{c|}{HeartLang} & {\ul{94.46}} & 0.19 & 69.60 & \multicolumn{1}{c|}{2.77} & 82.03 \\
\rowcolor[HTML]{EFEFEF}
\multicolumn{1}{c|}{SPOTR} &
$\mathbf{95.45}_{\textcolor{red}{\uparrow\,0.99}}$ & 0.39 &
$\mathbf{89.94}_{\textcolor{red}{\uparrow\,4.15}}$ & \multicolumn{1}{c|}{2.07} &
$\mathbf{92.70}_{\textcolor{red}{\uparrow\,10.06}}$ \\ \midrule

\rowcolor{purple!15}
\multicolumn{6}{c}{PPG} \\
\multicolumn{1}{c|}{\textbf{Dataset}} & \multicolumn{2}{c}{\textbf{PPG-Dalia}} & \multicolumn{2}{c|}{\textbf{WESAD$_{\text{v}}$}} & \\
\multicolumn{1}{c|}{Model} & mean & std. & mean & \multicolumn{1}{c|}{std.} & \multirow{-2}{*}{Ave} \\ \midrule
\multicolumn{1}{c|}{MOMENT}    & 80.20 & 0.03 & {\ul{68.22}} & \multicolumn{1}{c|}{0.13} & 74.21 \\
\multicolumn{1}{c|}{PaPaGei}   & {\ul{82.41}} & 0.23 & 66.04 & \multicolumn{1}{c|}{2.81} & 74.23 \\
\multicolumn{1}{c|}{Pulse-PPG} & \textbf{84.14} & 0.48 & 67.43 & \multicolumn{1}{c|}{3.10} & \textbf{75.79} \\
\rowcolor[HTML]{EFEFEF}
\multicolumn{1}{c|}{SPOTR} &
$82.30_{\textcolor{green}{\downarrow\,1.84}}$ & 0.81 &
$\mathbf{68.87}_{\textcolor{red}{\uparrow\,0.65}}$ & \multicolumn{1}{c|}{1.97} &
$\underline{75.59}_{\textcolor{green}{\downarrow\,0.20}}$ \\
\bottomrule
\end{tabular}%
}
\end{table}

\subsection{Efficiency}

In the efficiency evaluation, we compare SPOTR against the general time-series foundation model MOMENT across 17 downstream datasets, reporting the mean results over four metrics: Latency (p50/p95), Throughput, and Peak GPU Memory. As shown in Table \ref{table5_efficiency}, SPOTR achieves substantially better inference efficiency. Specifically, the average p50 and p95 latency are reduced from 32.61 ms and 35.48 ms to 7.15 ms and 7.69 ms, corresponding to 78.1\% and 78.3\% lower latency, respectively. Meanwhile, throughput increases from 45.66 to 139.84 samples/s, yielding a 206.3\% improvement. SPOTR also reduces peak GPU memory consumption from 532.09 MB to 256.39 MB (51.8\% lower). These results demonstrate that SPOTR delivers markedly faster and more memory-efficient inference, highlighting its practicality for resource-constrained deployment. More details are provided in the Appendix. 

\begin{table}[H]
\caption{\textbf{Efficiency comparison.}}
\label{table5_efficiency}
\centering
\resizebox{0.9\columnwidth}{!}{
\begin{tabular}{lcccc}
\toprule
& p50 Lat.$\downarrow$ & p95 Lat.$\downarrow$ & Thrpt.$\uparrow$ & Mem.(MB)$\downarrow$ \\
\midrule
MOMENT & 32.61 & 35.48 & 45.66  & 532.09 \\
\rowcolor[HTML]{EFEFEF}
SPOTR  & \textbf{7.15} & \textbf{7.69} & \textbf{139.84} & \textbf{256.39} \\
\midrule
$\Delta$(\%)
& -78.1\%
& -78.3\%
& +206.3\%
& -51.8\% \\
\bottomrule
\end{tabular}
}
\end{table}

\subsection{Scaling Model Size}

\begin{figure}[htb]
\begin{center}
\centerline{\includegraphics[width=\columnwidth]{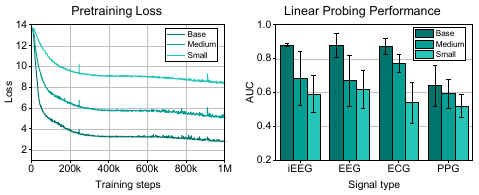}}
\caption{\textbf{Scaling behavior of SPOTR with model size}. Left: pretraining loss curves for different model sizes. Right: linear-probing AUC across four modalities improves consistently with model size.}
\label{fig4_scaling_law}
\end{center}
\vspace{-0.5cm}
\end{figure}

As shown in Fig.~\ref{fig4_scaling_law}, increasing the model size from SPOTR-Small (3.69M parameters) to SPOTR-Medium (13.93M) and SPOTR-Base (62.47M) leads to a clear scaling trend in pretraining. Larger models drop the loss faster early on and settle at a lower value, suggesting stronger capacity and easier optimization. The same pattern appears in linear probing. SPOTR-Base gives the best AUC on iEEG, EEG, ECG, and PPG, followed by SPOTR-Medium and then SPOTR-Small. Overall, scaling up the model improves both reconstruction training and representation transfer across signal types, which supports further scaling of model and data.

\section{Conclusion}

In this paper, we presented SPOTR, a universal self-supervised learning framework for physiological signals that learns generalizable representations across modalities via a compress–reconstruct pretraining scheme. By compressing each input into one representation and reconstructing the signal only from this bottleneck, SPOTR discourages shortcut learning and promotes compact global representations. SPOTR further improves efficiency by shortening the token sequence before modeling. Extensive experiments demonstrate consistent gains in cross-modality generalization, with particularly strong improvements under linear probing, supporting lightweight and practical adaptation in real-world clinical scenarios, while reducing inference latency and peak GPU memory compared with a representative general-purpose time-series foundation model.

\section*{Acknowledgments}
This work has been supported by Guangdong S\&T Program (2025B0101140001, 2026B0101070006), Guangdong Provincial Key Laboratory of In-Memory Computing Chips (2024B1212020002), Shenzhen Science and Technology Program (ZDCY20250901103401002, JCYJ20241202125907011), and Beijing Natural Science Foundation (L234026, L257010). This work has been supported by the New Cornerstone Science Foundation and Financial Support for Outstanding scientific and technological innovation Talents Training Fund in Shenzhen.

\bibliographystyle{named}
\bibliography{ijcai26}

\newpage

\typeout{IJCAI--ECAI 26 Instructions for Authors}


\pdfpagewidth=8.5in
\pdfpageheight=11in




\linenumbers

\urlstyle{same}







\pdfinfo{
/TemplateVersion (IJCAI.2026.0)
}

\title{IJCAI--ECAI 26 Formatting Instructions}

\appendix
\onecolumn

\section*{Appendix}

\section{More Details on Experimental Setup}

\subsection{Baselines}

\textbf{MOMENT} \cite{moment} is a foundation model for multivariate time-series signals across domains (e.g., healthcare, engineering, finance). It segments each series into fixed-length patch tokens and pretrains via masked time-series prediction, reconstructing masked patches to learn transferable representations for forecasting, classification, anomaly detection, and imputation. The raw code is available at \url{https://github.com/moment-timeseries-foundation-model/moment}.

\textbf{BIOT} \cite{biot} is a generic transformer for biosignals (EEG, ECG, and wearable sensors). It tokenizes each channel into fixed-length segments, flattens them into a unified biosignal sentence, and adds channel and relative position embeddings before a linear-complexity transformer. For self-supervision, it uses channel/token dropout and contrastive loss. The raw code is available at \url{https://github.com/ycq091044/BIOT}.

\textbf{LaBraM} \cite{labram} is a foundation model for EEG signals. It segments EEG into channel patches, trains a vector-quantized neural tokenizer via neural spectrum prediction, then pretrains a Transformer by masked code prediction (BEiT/MAE-style) to learn transferable EEG representations across datasets and tasks. The raw code is available at \url{https://github.com/935963004/LaBraM}.

\textbf{CBraMod} \cite{cbramod} is a foundation model for EEG signals. It pretrains with patch-based masked EEG reconstruction using a time-frequency patch encoder, then applies a criss-cross transformer with separate spatial/temporal attentions and asymmetric conditional positional encoding to adapt across diverse channel layouts. The raw code is available at \url{https://github.com/wjq-learning/CBraMod}.

\textbf{CSBrain} \cite{csbrain} is a brain foundation model for EEG signals. It introduces cross-scale spatiotemporal tokenization to aggregate multi-scale features within local temporal windows and anatomical brain regions into scale-aware tokens, then applies structured sparse attention to model cross-window/region dependencies efficiently while suppressing spurious correlations. The raw code is available at \url{https://github.com/yuchen2199/CSBrain}.

\textbf{HeartLang} \cite{heartlang} is a self-supervised foundation model for electrocardiogram (ECG) signals. It treats heartbeats as “words” and rhythms as “sentences”, using a QRS-Tokenizer to convert raw ECG into ECG sentences, then pretrains a transformer backbone (ST-ECGFormer) to learn form- and rhythm-level representations. The raw code is available at \url{https://github.com/PKUDigitalHealth/HeartLang}.

\textbf{ST-MEM} \cite{stmem} is a self-supervised foundation model for 12-lead electrocardiogram (ECG) signals. It performs spatio-temporal masked modeling, patchifying ECG across leads and time, then reconstructing masked regions to capture lead-wise (spatial) and temporal relationships for transferable ECG representations. The raw code is available at \url{https://github.com/vuno/ST-MEM}.

\textbf{PaPaGei} \cite{papagei} is an open foundation model for photoplethysmography (PPG) signals. It is pretrained on 57k+ hours of unlabeled PPG segments and uses two SSL objectives: participant-aware contrastive learning and morphology-aware learning that matches segments with similar waveform metrics (sVRI/IPA/SQI) via specialized heads. The raw code is available at \url{https://github.com/Nokia-Bell-Labs/papagei-foundation-model}.

\textbf{Pulse-PPG} \cite{pulseppg} is a foundation model for wearable photoplethysmography (PPG) signals. It is pretrained on raw, uncurated smartwatch PPG from a 100-day field study (120 participants) and learns representations via a learnable motif-based distance function plus relative contrastive learning to enforce correct similarity ordering under noise. The raw code is available at \url{https://github.com/maxxu05/pulseppg}.

\subsection{Pretraining Datasets}

\subsubsection{EEG Datasets}

\textbf{Alzheimer’s Disease Dataset} \cite{alzheimer}. This resting-state (eyes-closed) scalp EEG dataset includes 88 participants: Alzheimer’s disease (AD, n=36), frontotemporal dementia (FTD, n=23), and cognitively normal controls (CN, n=29). 

\textbf{SleepEDF} \cite{sleepedf}. The dataset contains 197 full-night PSG recordings, including EEG, EOG, submental EMG, and manually labeled sleep-stage annotations (with annotated sleep events).

\textbf{Sleep Heart Health Study (SHHS)} \cite{shhs}. The Sleep Heart Health Study (SHHS) is a multicenter cohort designed to study links between sleep-disordered breathing and cardiovascular outcomes. SHHS-1 collected baseline full-night PSG from 6,441 adults, and SHHS-2 repeated PSG in 3,295 participants, with outcomes followed to $\sim$2010. PSG includes EEG (C3/A2, C4/A1; 125 Hz) along with standard EOG/EMG and respiratory/$SpO_2$ channels. In this work, we use only EEG in SHHS-1.

\textbf{CHB-MIT Scalp EEG Dataset} \cite{chbmit}. The CHB-MIT Scalp EEG Dataset provides long-term scalp EEG from 22 pediatric patients with intractable epilepsy at Boston Children’s Hospital, collected during medication withdrawal. It includes 686 sessions with 182 annotated seizures.

\textbf{TDBRAIN Database} \cite{tdbrain}. The TDBRAIN database provides resting-state scalp EEG from 1,274 participants collected between 2001 and 2021, spanning multiple psychiatric conditions (e.g., MDD=426, ADHD=271, SMC=119, OCD=75) as well as healthy controls (47) and undiagnosed cases (255). For each participant, one 2-min eyes-open and one 2-min eyes-closed segment were selected. 

\textbf{NMT Scalp EEG Dataset} \cite{nmt}. The NUST-Military Hospital–TUKL EEG dataset contains 2,417 EEG recordings from unique participants ($\sim$625 hours total), each labeled by expert neurologists as normal or abnormal.

\subsubsection{iEEG Datasets}

\textbf{SWEC-ETHZ Dataset} \cite{swecethz}. Released by the Sleep–Wake–Epilepsy Center (SWEC, Inselspital Bern) with ETH Zurich, this dataset provides continuous presurgical iEEG from 18 patients with pharmacoresistant epilepsy. It contains $>2,656$ hours of recordings and 116 expert-annotated seizures. 

\textbf{Epilepsy-iEEG-Multicenter Dataset} \cite{multicenterieeg}. This multicenter epilepsy dataset includes iEEG and scalp EEG from 100 patients with refractory epilepsy, covering depth/strip/surface electrodes (typically 2,000 Hz) and both pre-surgical and post-surgical recordings with seizure annotations and surgical/outcome metadata. In this study, we use only iEEG.

\subsubsection{ECG Datasets}

\textbf{MIMIC-IV-ECG} \cite{mimic}. MIMIC-IV-ECG provides $\sim$800,000 diagnostic 12-lead (10 s) ECGs from $\sim$160,000 patients, linked to the MIMIC-IV clinical database. Recordings are sampled at 500 Hz and include automated measurements and, when available, cardiologist-confirmed reports. 

\textbf{Chapman-Shaoxing ECG Database} \cite{shaoxing}. Chapman University and Shaoxing People’s Hospital jointly released this dataset, which contains 10,646 patients’ 10-second 12-lead ECG recordings sampled at 500 Hz. Each recording is annotated by cardiologists with rhythm labels (11 classes) and cardiovascular pathology categories (67 classes).

\textbf{Shandong Provincial Hospital ECG Database} \cite{shandong}. This dataset contains 25,770 multi-label 12-lead ECG recordings from 24,666 patients. Signals are sampled at 500 Hz with durations of 10–60s. Diagnostic statements follow AHA/ACC/HRS standards, covering 44 primary diagnoses and 15 modifiers.

\textbf{Lobachevsky University ECG Database (LUDB)} \cite{ludb}. LUDB provides 10-second, 12-lead ECGs from 200 subjects recorded at 500 Hz. Cardiologists manually annotated P/QRS/T wave boundaries and peaks ($\sim$58,429 beats) and provided diagnostic labels; subjects include healthy controls and patients with diverse cardiac conditions (including pacemakers).

\textbf{PTB Diagnostic ECG Database} \cite{ptbdiagnostic}. This dataset contains 549 high-resolution ECG recordings from 294 subjects, including healthy volunteers and patients with various cardiac diseases (e.g., MI, cardiomyopathy, conduction disorders, arrhythmia, hypertrophy, valvular disease, myocarditis). Each record provides 15 leads (12 standard + Frank XYZ) sampled at 1,000 Hz (16-bit).

\textbf{St. Petersburg INCART Database} \cite{stpeter}. This database provides 75 half-hour 12-lead Holter ECG recordings from 32 patients with coronary artery disease, sampled at 257 Hz, with $>175,000$ annotated beats. The cohort covers multiple confirmed cardiac conditions (e.g., myocardial ischemia, hypertension, atrial fibrillation, sick sinus syndrome, ventricular ectopy).

\textbf{Georgia 12-Lead ECG Challenge Dataset} \cite{geogia}. Provided by Emory University as part of the PhysioNet 2020 Challenge, this Southeastern U.S. cohort provides 10,344 12-lead ECG recordings (5–10 s, 500 Hz) with demographics and diagnostic labels. As the validation and test sets are not public, we use only the released training set.

\textbf{CODE-15\% Dataset} \cite{code15}. This dataset includes 345,779 12-lead ECGs from 233,770 patients collected by the Telehealth Network of Minas Gerais (2010–2016), covering $\sim$15\% of the full CODE cohort. Metadata provide age/sex and structured labels for normal/abnormal status and multiple rhythm or conduction abnormalities.

\subsubsection{PPG Datasets}

\textbf{Pulse Transit Time PPG Dataset} \cite{pulseppg}. Provided by the University of Sydney, this dataset includes multimodal physiological recordings from 22 healthy participants during sitting, walking, and running. It provides dual-site, multi-wavelength PPG (660/537/880 nm; 1,000 Hz) synchronized with IMU (500 Hz), ECG (500 Hz), and blood pressure measurements.

\textbf{BUT PPG Dataset} \cite{butppg}. Developed by Brno University of Technology, this dataset provides 3,888 smartphone-camera PPG recordings (10 s, 30 Hz) from 50 participants under rest and motion conditions (e.g., walking, speaking/coughing, illumination changes, finger/earlobe motion).

\textbf{Pediatric SDB Dataset} \cite{sdb}. This dataset contains PPG, $SpO_2$, and heart rate signals from children in awake state. It is designed for preliminary screening of sleep-disordered breathing (e.g., obstructive sleep apnea, OSAS), targeting moderate-to-severe cases. We used it for binary classification of high-risk vs. low-risk OSAS. 

\subsubsection{Multimodal Dataset VitalDB}

\textbf{VitalDB} \cite{vitaldb} contains intraoperative vital sign recordings from 6,388 non-cardiac surgical patients at Seoul National University Hospital (2016–2017). The dataset integrates high-resolution waveforms and perioperative clinical data, including 557,622 signal and numeric tracks from 196 monitoring parameters (e.g., ECG, EEG, BP, $SpO_2$, temperature, bispectral index, BIS), 74 perioperative clinical variables, and 34 laboratory time-series. Waveforms were sampled at 500 Hz, while numeric data were recorded at 1–7 s intervals. For pretraining, we used EEG and PPG signals.

\subsection{Downstream Datasets}

\subsubsection{EEG Datasets}

\textbf{ISRUC-Sleep dataset} \cite{isruc}. ISRUC-Sleep includes three cohorts (100 subjects with a single PSG, 8 with two PSGs, and 10 healthy controls) with expert annotations based on EEG/EOG/EMG and respiratory signals. We use EEG from the 100-subject single-PSG cohort for 5-class sleep staging. 

\textbf{NTUH Anesthesia Depth EEG Dataset} \cite{ntuh-bis}. This dataset provides EEG from 24 patients under general anesthesia ($\sim$50 hours, 128 Hz), spanning consciousness states from wakefulness to deep anesthesia. Using the Bispectral Index (BIS), segments are labeled into five classes: Awake/Light Sedation (80–100), Moderate Sedation (60–80), General Anesthesia (40–60), Deep Hypnosis (20–40), and Burst Suppression (0–20). 

\textbf{MDD Patients and Healthy Controls EEG Data} \cite{mdd}. The MDD dataset was designed to study EEG differences between patients with major depressive disorder (MDD) and healthy controls (HC). It contains resting-state EEG from 34 MDD patients and 30 healthy controls, sampled at 256 Hz. We used this dataset for binary classification of MDD vs. healthy controls. 

\textbf{Schizophrenia-28 EEG (SCZ)} \cite{schizo}. This dataset includes standardized resting-state EEG recordings from 14 patients with paranoid schizophrenia and 14 age-matched healthy controls. We used this dataset for binary classification of schizophrenia vs. healthy controls.

\textbf{Siena Scalp EEG Database} \cite{siena}. This dataset contains multi-channel EEG from 14 adult epilepsy patients (9 male, ages 25–71; 5 female, ages 20–58), sampled at 512 Hz using electrodes placed according to the international 10–20 system. The dataset includes more than 128 hours of recordings with 47 seizures, all precisely annotated by experts with onset and offset labels. We used this dataset for binary seizure detection. 

\subsubsection{iEEG Datasets}

\textbf{FNUSA Dataset} \cite{mayo}. The FNUSA dataset contains interictal resting-state iEEG from 14 drug-resistant epilepsy patients undergoing presurgical monitoring. Recordings were acquired with depth electrodes (25 kHz), low-pass filtered at 2 kHz, and downsampled to 5 kHz. We use it for binary detection of epileptiform discharges. 

\textbf{Mayo Dataset} \cite{mayo}. The Mayo dataset contains iEEG from 25 drug-resistant epilepsy patients recorded during the first night after electrode implantation. Signals were acquired at 32 kHz, then low-pass filtered at 1 kHz and downsampled to 5 kHz using depth/grid/strip electrodes. Segments are annotated as physiological activity, pathological activity (e.g., spikes/HFOs), line noise (60 Hz), or other non-cerebral artifacts. We use it for binary detection of epileptiform discharges.

\subsubsection{ECG Datasets}

\textbf{CPSC-2018} \cite{cpsc2018}. The CPSC-2018 dataset contains 9,831 ECG recordings from 9,458 patients, collected in clinical environments at 11 hospitals across China. Recordings range from 7 to 60 minutes and cover 9 ECG classes: normal and 8 abnormalities (AF, I-AVB, LBBB, RBBB, PAC, PVC, STD, STE). The dataset is widely used for multi-label ECG abnormality detection. We used it for multi-label cardiac disease classification. 

\textbf{PTB-XL} \cite{ptbxl}. PTB-XL contains 21,799 10-second 12-lead ECG recordings from 18,869 patients (0–95 years, 52\% male, 48\% female). Each record was annotated by up to two cardiologists using 71 SCP-ECG statements across diagnostic, morphological, and rhythm categories, consolidated into 5 superclasses and 23 subclasses. The dataset provides standardized train–test splits and supports multiple tasks, including multi-label classification, rhythm/morphology classification, and signal quality assessment. We followed common classification schemes at superclass, subclass, rhythm, and morphology levels \cite{merl}. 

\textbf{European ST-T Database} \cite{europe_st-t}. This dataset was developed to evaluate algorithms for ST–T wave analysis. It includes 90 2-hour 2-lead ECG recordings from 79 subjects (70 men, 8 women, 1 unknown), covering $\sim$300 hours of data, sampled at 250 Hz with 12-bit resolution. Two cardiologists annotated each beat and ST–T event, including onset, extreme point, and offset, yielding $>800,000$ annotations. We used this dataset for heartbeat classification. 

\subsubsection{PPG Datasets}

\textbf{PPG-BP} \cite{ppg-bp}. This dataset contains 657 short PPG recordings from 219 Chinese subjects (ages 20–89), each paired with synchronous systolic (SBP) and diastolic (DBP) blood pressure values, covering normal and hypertensive states. We used it for hypertension classification.

\textbf{Cuff-Less BP Estimation Dataset} \cite{clbp}. This dataset includes 4,254 segments of PPG, ECG, and ABP signals from the MIMIC-II database, with blood pressure measured using cuff-less devices. We used it for hypertension classification. 

\textbf{PPG-DaLiA} \cite{ppgdalia}. The PPG-DaLiA (Photoplethysmography Dataset for Daily Life Activities) dataset contains up to 36 hours of multimodal recordings from 15 healthy subjects during daily-life activities (e.g., walking, cycling, driving). We use this dataset for human activity recognition.

\textbf{WESAD} \cite{wesad}. WESAD is a wearable stress and affect dataset collected from 15 subjects in a controlled lab protocol with baseline, stress, and amusement conditions. Recordings are multimodal and come from both a chest-worn and a wrist-worn device, covering physiological signals (e.g., ECG, EDA, EMG, respiration, temperature/BVP) and 3-axis acceleration, together with self-reports. In our experiments, we use it for binary emotion classification based on valence and arousal from self-reports.

\section{Additional Results}

\subsection{Inference Efficiency}

In the efficiency benchmark, we compare SPOTR with MOMENT on 17 downstream datasets (Table~\ref{tab:efficiency}) using a unified inference-only protocol. For each dataset, we instantiate a fresh model with the corresponding input channel configuration and run forward-only inference on randomly generated inputs. On GPU, latency is measured using CUDA events with 50 warm-up iterations followed by 200 timed iterations, from which we report p50 and p95. Throughput is computed as $1000/\text{p50(ms)}$ (samples\,s$^{-1}$), and peak GPU memory is obtained from the maximum allocated memory during the timed phase after resetting memory statistics. Each setting is repeated five times, and we report the median across repeats, while clearing CUDA caches and releasing the model between datasets to avoid cross-dataset interference.

\begin{table}[htb]
\centering
\caption{Inference efficiency comparison between MOMENT and SPOTR across datasets. SPOTR achieves lower latency and peak memory while increasing throughput.}
\footnotesize
\setlength{\tabcolsep}{4.2pt}
\renewcommand{\arraystretch}{1.12}
\begin{tabular}{
    l
    S[table-format=2.2] S[table-format=1.2]
    S[table-format=2.2] S[table-format=1.2]
    S[table-format=3.2] S[table-format=3.2]
    S[table-format=3.2] S[table-format=3.2]
}
\toprule
\textbf{Dataset} &
\multicolumn{2}{c}{\textbf{Latency (ms, p50)}} &
\multicolumn{2}{c}{\textbf{Latency (ms, p95)}} &
\multicolumn{2}{c}{\textbf{Throughput (samples/s)}} &
\multicolumn{2}{c}{\textbf{Peak GPU Memory (MB)}} \\
\cmidrule(lr){2-3}\cmidrule(lr){4-5}\cmidrule(lr){6-7}\cmidrule(lr){8-9}
& {MOMENT} & {\textbf{SPOTR}} & {MOMENT} & {\textbf{SPOTR}} & {MOMENT} & {\textbf{SPOTR}} & {MOMENT} & {\textbf{SPOTR}} \\
\midrule
ISRUC        & 46.01 & 7.50 & 47.95 & 7.99 & 21.73 & 133.36 & 645.21 & 257.14 \\
Siena        & 42.95 & 7.15 & 45.28 & 7.68 & 23.29 & 139.90 & 567.12 & 259.70 \\
NTUH-BIS     & 29.57 & 7.46 & 30.40 & 7.95 & 33.82 & 134.13 & 544.58 & 254.45 \\
MDD          & 37.49 & 7.12 & 43.52 & 7.65 & 26.67 & 140.46 & 547.23 & 258.69 \\
Schizo28     & 79.65 & 7.25 & 81.70 & 7.75 & 12.55 & 137.85 & 697.22 & 263.76 \\
Mayo         & 11.82 & 6.94 & 13.37 & 7.48 & 84.59 & 144.01 & 448.66 & 252.98 \\
FNUSA        & 12.14 & 7.01 & 13.35 & 7.48 & 82.38 & 142.74 & 448.66 & 252.98 \\
CPSC2018     & 53.14 & 7.06 & 56.66 & 7.68 & 18.82 & 141.63 & 603.78 & 259.87 \\
PTBXL-sub    & 51.37 & 7.09 & 56.45 & 7.69 & 19.47 & 140.98 & 603.78 & 259.87 \\
PTBXL-super  & 51.00 & 7.15 & 56.10 & 7.69 & 19.61 & 139.90 & 603.78 & 259.87 \\
PTBXL-rhythm & 51.02 & 7.12 & 56.25 & 7.68 & 19.60 & 140.45 & 603.78 & 259.87 \\
PPG-BP       & 15.90 & 7.27 & 18.78 & 7.76 & 62.87 & 137.56 & 455.06 & 253.31 \\
PPG-Dalia    & 12.48 & 7.21 & 13.72 & 7.74 & 80.13 & 138.75 & 455.75 & 253.26 \\
WESAD$_v$      & 15.92 & 7.10 & 18.75 & 7.59 & 62.82 & 140.90 & 457.79 & 253.36 \\
WESAD$_a$      & 15.94 & 7.07 & 18.77 & 7.59 & 62.72 & 141.43 & 457.79 & 253.36 \\
Europe ST-T  & 12.19 & 7.11 & 13.41 & 7.67 & 82.01 & 140.67 & 447.55 & 252.81 \\
CLBP         & 15.85 & 7.00 & 18.71 & 7.59 & 63.10 & 142.83 & 457.79 & 253.36 \\
\midrule
\textbf{Average} & \textbf{32.61} & \textbf{7.15} & \textbf{35.48} & \textbf{7.69} & \textbf{45.66} & \textbf{139.84} & \textbf{532.09} & \textbf{256.39} \\
\bottomrule

\end{tabular}
\label{tab:efficiency}
\end{table}

\subsection{Scaling Model Size}

To study how model scaling affects both design choices and representation transfer, we report the detailed configurations of SPOTR at different sizes (Small/Medium/Base) in Table \ref{tab:spotr_scales}. We then evaluate these scaled variants under a unified linear-probing protocol across downstream datasets from four signal modalities: iEEG (Table \ref{scale_ieeg}), EEG (Table \ref{scale_eeg}), ECG (Table \ref{scale_ecg}), and PPG (Table \ref{scale_ppg}). Together, these tables provide a complete view of SPOTR’s scaling setup and its corresponding transfer performance across heterogeneous physiological signals.

\begin{table}[H]
\centering
\caption{Model configurations of SPOTR at different scales.}
\small
\setlength{\tabcolsep}{6pt}
\renewcommand{\arraystretch}{1.1}
\begin{tabular}{lccccc}
\toprule
\textbf{Scale} & \textbf{Encoder Params} & \textbf{Encoder Dim} & \textbf{\# Encoder Layers} & \textbf{Decoder Dim} & \textbf{\# Decoder Layers} \\
\midrule
Small  & 3.69M  & 128 & 4  & 64  & 2 \\
Medium & 13.93M & 256 & 8  & 128 & 4 \\
Base   & 62.47M & 512 & 12 & 256 & 6 \\
\bottomrule
\end{tabular}
\label{tab:spotr_scales}
\end{table}

\begin{table}[H]
\centering
\caption{Linear-probing performance of scaled SPOTR models on iEEG downstream datasets.}
\label{scale_ieeg}
\resizebox{0.45\textwidth}{!}{%
\begin{tabular}{c|cccc|c}
\hline
\cellcolor[HTML]{FFFFFF}\textbf{Dataset} & \multicolumn{2}{c}{\textbf{Mayo}}                                     & \multicolumn{2}{c|}{\textbf{FNUSA}}                                   &                       \\
Model                                    & mean                                   & std                          & mean                                   & std                          & \multirow{-2}{*}{Ave} \\ \hline
SPOTR-Small                              & 67.11                                  & 0.06                         & 51.55                                  & 1.56                         & 59.33                 \\
SPOTR-Medium                             & 79.79                                  & 0.11                         & 57.42                                  & 4.92                         & 68.61                 \\
\rowcolor[HTML]{EFEFEF} 
SPOTR-Base       & \textbf{88.92} & 0.06 & \textbf{87.52} & 0.16 & \textbf{88.22}        \\ \hline
\end{tabular}%
}
\end{table}

\begin{table}[H]
\centering
\caption{Linear-probing performance of scaled SPOTR models on EEG downstream datasets.}
\label{scale_eeg}
\resizebox{0.8\textwidth}{!}{%
\begin{tabular}{c|cccccccccc|c}
\hline
\cellcolor[HTML]{FFFFFF}\textbf{Dataset} & \multicolumn{2}{c}{\textbf{ISRUC}} & \multicolumn{2}{c}{\textbf{Siena}} & \multicolumn{2}{c}{\textbf{NTUH}} & \multicolumn{2}{c}{\textbf{MDD}} & \multicolumn{2}{c|}{\textbf{Schizo28}} &                       \\
Model                                    & mean                  & std        & mean                  & std        & mean                 & std        & mean                 & std       & mean                    & std          & \multirow{-2}{*}{Ave} \\ \hline
SPOTR-Small                              & 64.66                 & 0.08       & 55.91                 & 3.05       & 49.33                & 0.55       & 78.94                & 0.20      & 59.90                   & 1.56         & 61.75                 \\
SPOTR-Medium                             & 76.20                 & 0.06       & 46.45                 & 1.59       & 58.37                & 0.87       & 84.70                & 0.07      & 70.26                   & 1.32         & 67.19                 \\
\rowcolor[HTML]{EFEFEF} 
SPOTR-Base                               & \textbf{92.83}        & 0.03       & \textbf{76.87}        & 0.90       & \textbf{87.69}       & 0.13       & \textbf{95.31}       & 0.12      & \textbf{87.79}          & 8.87         & \textbf{88.10}        \\ \hline
\end{tabular}%
}
\end{table}

\begin{table}[H]
\centering
\caption{Linear-probing performance of scaled SPOTR models on ECG downstream datasets.}
\label{scale_ecg}
\resizebox{0.8\textwidth}{!}{%
\begin{tabular}{c|cccccccccc|c}
\hline
\cellcolor[HTML]{FFFFFF}\textbf{Dataset} & \multicolumn{2}{c}{\textbf{CPSC2018}} & \multicolumn{2}{c}{\textbf{PTBXL-sub}} & \multicolumn{2}{c}{\textbf{PTBXL-super}} & \multicolumn{2}{c}{\textbf{PTBXL-rhythm}} & \multicolumn{2}{c|}{\textbf{Europe ST-T}} &                       \\
Model                                    & mean                   & std          & mean                    & std          & mean                     & std           & mean                     & std            & mean                     & std            & \multirow{-2}{*}{Ave} \\ \hline
SPOTR-Small                              & 50.66                  & 1.12         & 48.75                   & 0.63         & 51.29                    & 0.85          & 45.13                    & 0.48           & 75.08                    & 0.11           & 54.18                 \\
SPOTR-Medium                             & 72.29                  & 0.86         & 80.74                   & 0.08         & 83.88                    & 0.05          & 71.31                    & 0.22           & 77.99                    & 0.19           & 77.24                 \\
\rowcolor[HTML]{EFEFEF} 
SPOTR-Base                               & \textbf{92.33}         & 0.03         & \textbf{90.21}          & 0.02         & \textbf{89.10}           & 0.01          & \textbf{82.84}           & 0.04           & \textbf{82.47}           & 0.19           & \textbf{87.39}        \\ \hline
\end{tabular}%
}
\end{table}

\begin{table}[H]
\centering
\caption{Linear-probing performance of scaled SPOTR models on PPG downstream datasets.}
\label{scale_ppg}
\resizebox{0.8\textwidth}{!}{%
\begin{tabular}{c|cccccccccc|c}
\hline
\cellcolor[HTML]{FFFFFF}\textbf{Dataset} & \multicolumn{2}{c}{\textbf{PPG-BP}} & \multicolumn{2}{c}{\textbf{PPG-Dalia}} & \multicolumn{2}{c}{\textbf{WESAD$_v$}} & \multicolumn{2}{c}{\textbf{WESAD$_a$}} & \multicolumn{2}{c|}{\textbf{CLBP}} &                       \\
Model                                    & mean                  & std         & mean                    & std          & mean                   & std         & mean                   & std         & mean                  & std        & \multirow{-2}{*}{Ave} \\ \hline
SPOTR-Small                              & 40.69                 & 6.86        & 59.03                   & 0.17         & 53.41                  & 0.55        & 54.09                  & 0.40        & 52.89                 & 0.05       & 52.02                 \\
SPOTR-Medium                             & 50.13                 & 11.21       & 69.21                   & 0.15         & 67.66                  & 0.20        & 55.48                  & 1.07        & 54.33                 & 0.06       & 59.36                 \\
\rowcolor[HTML]{EFEFEF} 
SPOTR-Base                               & \textbf{51.99}        & 8.26        & \textbf{82.12}          & 0.05         & \textbf{69.03}         & 0.49        & \textbf{62.06}         & 0.19        & \textbf{55.52}        & 0.25       & \textbf{64.14}        \\ \hline
\end{tabular}%
}
\end{table}

\vspace{0.5cm}

\section{Compression-Reconstruction Visualizations}

We present qualitative reconstruction results of the proposed compression–reconstruction pretraining paradigm on four representative datasets—Mayo (iEEG), MDD (EEG), and CPSC2018 (ECG), and PPG-DaLiA (PPG)—to probe robustness under diverse modalities and acquisition conditions as well as cross-dataset generalization. 

The left column shows the original signals and the right column shows the reconstructed signals; for multi-channel recordings, we visualize the first four channels to keep the layout consistent across datasets. 

Across all four modalities, the reconstructions largely preserve global morphology and temporal organization, suggesting that the compressed latent captures physiologically meaningful structure rather than merely copying local neighborhoods. For iEEG (Mayo), where signals are broadband and channel-dependent, the reconstructions remain stable and preserve cross-channel trends (e.g., shared slow components and consistent relative dynamics), implying that the model learns transferable spatiotemporal dependencies rather than overfitting to a specific montage. In EEG (MDD), the reconstructed traces retain prominent rhythmic structure and salient transient events while slightly smoothing fine-grained noise, which is consistent with the desired effect of a bottleneck: prioritizing task-relevant patterns over idiosyncratic high-frequency artifacts. In ECG (CPSC2018), the reconstructed waveform maintains the characteristic beat-level profile and beat-to-beat rhythm, indicating that sharp transient components (e.g., QRS-like peaks) remain recoverable under strong compression. Finally, in PPG (PPG-DaLiA), the reconstructed pulses track the dominant pulse shape and periodicity, reflecting robustness to motion-related variability typical of daily-life acquisition.

Overall, these qualitative examples indicate that SPOTR can reconstruct signals with consistent global structure across modalities and datasets, while minor deviations mainly appear in fine-grained details, which is expected under strong compression.

\newpage

\begin{figure}[H]
  \centering
  \includegraphics[width=\linewidth]{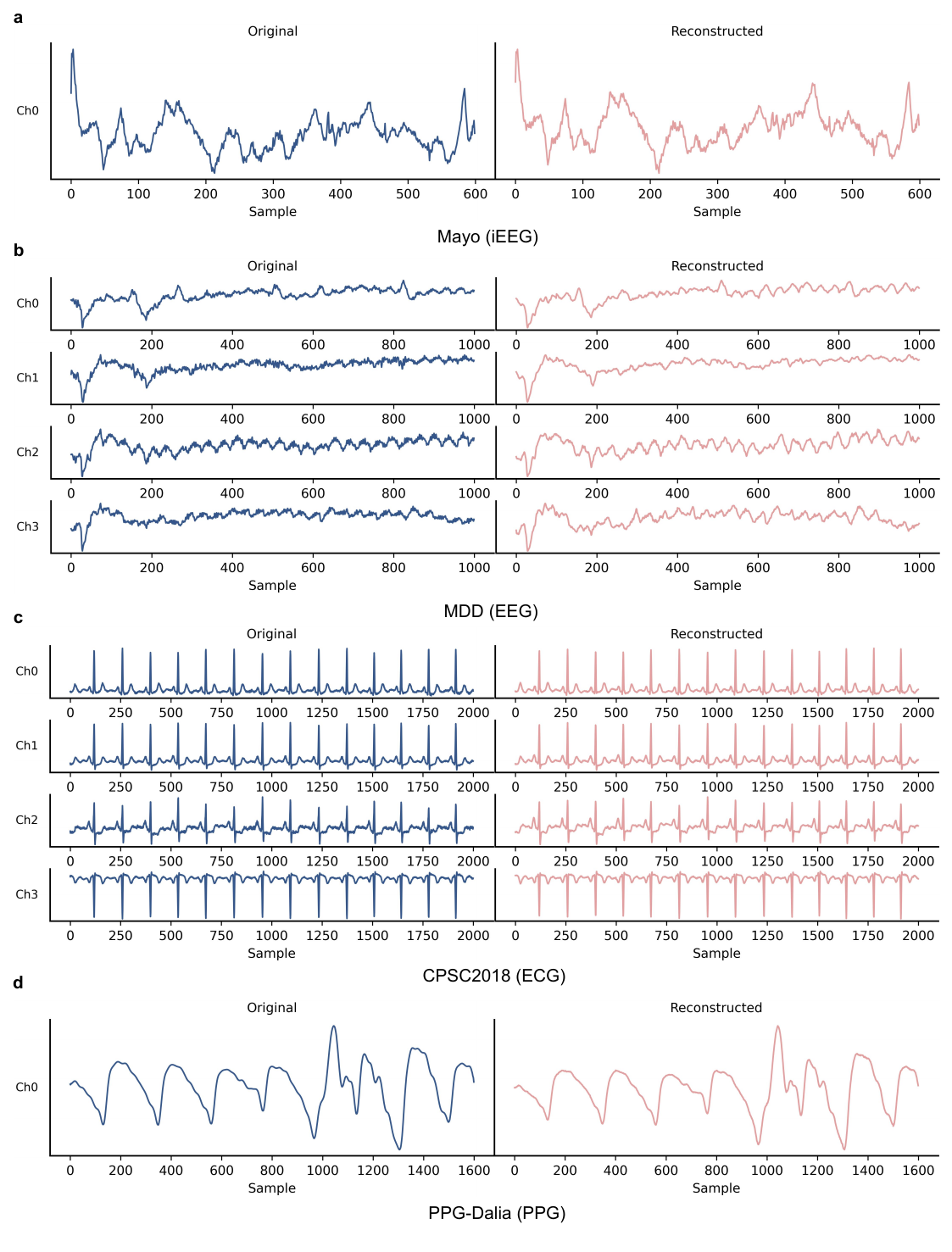}
  \caption{Qualitative reconstruction results across four datasets (ECG, EEG/iEEG, and PPG). For datasets with more than four channels, only the first four channels are visualized.}
  \label{fig:supp_recon}
\end{figure}

\newpage

\section{Visualization of Attention Patterns}

\begin{figure}[H]
  \centering
  \includegraphics[width=0.9\linewidth]{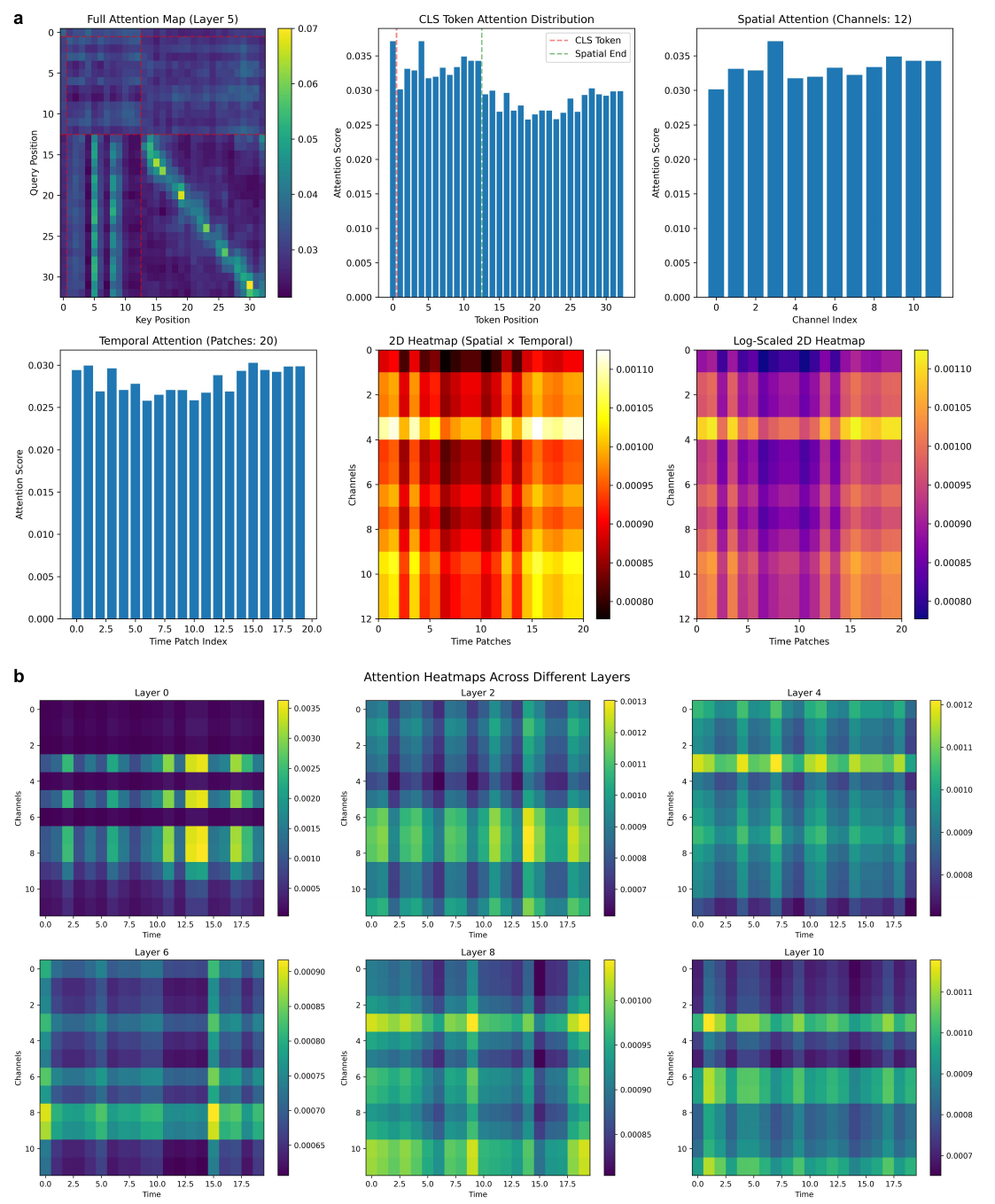}
  \caption{\textbf{Temporal--spatial attention visualization on CPSC2018.}
  (a) Detailed attention analysis for a representative layer (Layer 5). We show the full attention map (query vs.\ key positions), together with the attention distribution associated with the CLS token, and the decomposed spatial and temporal attentions over channels (12) and time patches (20), respectively. The resulting 2D attention heatmap (channels $\times$ time patches) is further visualized in both linear and log scales to highlight subtle yet consistent structures.
  (b) Attention heatmaps across different layers (Layers 0/2/4/6/8/10), all presented in the same temporal--spatial form (channels $\times$ time patches), illustrating how the model's attention allocation evolves with depth.}
  \label{fig:attn_cpsc2018}
\end{figure}

\begin{figure}[H]
  \centering
  \includegraphics[width=0.9\linewidth]{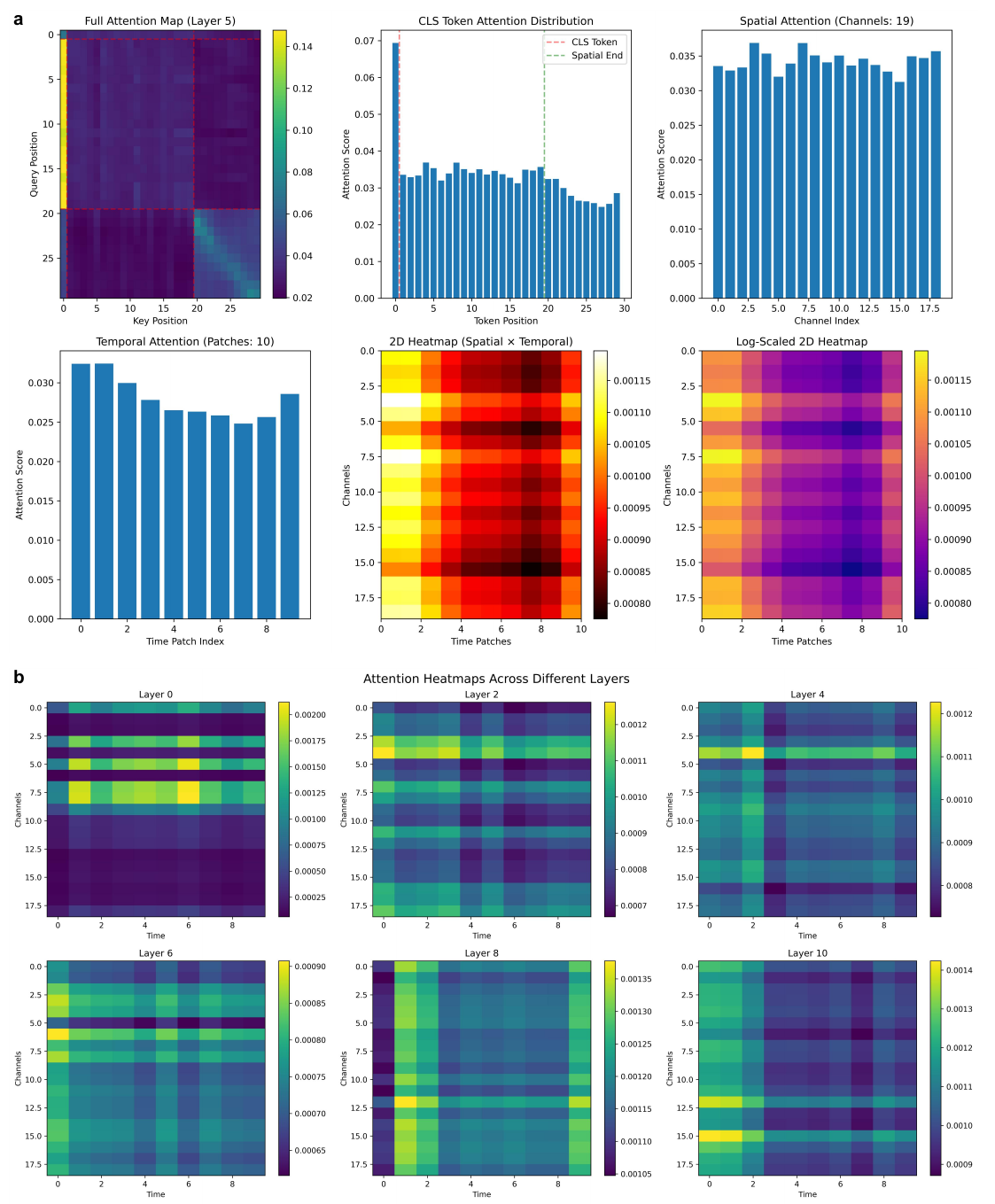}
  \caption{\textbf{Temporal--spatial attention visualization on the MDD dataset.}
  (a) Detailed attention analysis for a representative layer (Layer 5). We visualize the full attention map (query vs.\ key positions), the attention distribution associated with the CLS token, and the decomposed spatial and temporal attentions over channels (19) and time patches (10), respectively. The resulting two-dimensional attention heatmap (channels $\times$ time patches) is further illustrated in both linear and log scales to reveal fine-grained yet stable attention patterns.
  (b) Temporal--spatial attention heatmaps across different layers (Layers 0/2/4/6/8/10), highlighting the evolution of attention allocation over channels and time as network depth increases.}
  \label{fig:attn_mdd}
\end{figure}



\end{document}